\documentclass{article}

\usepackage{microtype}
\usepackage{graphicx}
\usepackage{subcaption}
\usepackage{booktabs} 
\usepackage{enumitem}
\usepackage{hyperref}

\usepackage[accepted]{icml2026}

\usepackage{amsmath}
\usepackage{amssymb}
\usepackage{mathtools}
\usepackage{amsthm}

\usepackage[capitalize,noabbrev]{cleveref}

\theoremstyle{plain}

\theoremstyle{definition}

\theoremstyle{remark}

\usepackage[textsize=tiny]{todonotes}

\icmltitlerunning{OpenMic: A Multi-Agent-Based Stand-Up Comedy Generation System}

\begin{document}

\twocolumn[
  \icmltitle{OpenMic: A Multi-Agent-Based Stand-Up Comedy Generation System}



  \icmlsetsymbol{equal}{*}

  \begin{icmlauthorlist}
    \icmlauthor{Yuyang Wu}{sch}
    \icmlauthor{Hanzhong Cao}{sch}
    \icmlauthor{Jianhao Chen}{sch}
    \icmlauthor{Yufei Li}{sch}
  \end{icmlauthorlist}
  
  \icmlaffiliation{sch}{School of Electronics Engineering and Computer Science, Peking University}

  \icmlcorrespondingauthor{Yuyang Wu}{wuyuyang@stu.pku.edu.cn}

  \icmlkeywords{CoRE}

  \vskip 0.3in
]



\printAffiliationsAndNotice{}  

\begin{abstract}
Chinese stand-up comedy generation goes beyond plain text generation, requiring culturally grounded humor, precise timing, stage-performance cues, and implicit multi-step reasoning. Moreover, commonly used Chinese humor datasets are often better suited for humor understanding and evaluation than for long-form stand-up generation, making direct supervision misaligned with the target task.
To address these challenges, we present \textbf{OpenMic}, an end-to-end multi-agent system built on AutoGen that transforms a user-provided life topic into a 3–5 minute Chinese stand-up performance and further produces a narrated comedy video. OpenMic orchestrates multiple specialized agents in a multi-round iterative loop—planning to jointly optimize humor, timing, and performability.
To mitigate the dataset–task mismatch, we augment generation with retrieval-augmented generation (RAG) for material grounding and idea expansion, and we fine-tune a dedicated JokeWriter to better internalize stand-up-specific setup–punchline structures and long-range callbacks. 
\end{abstract}

\section{Introduction}

Artificial intelligence has made rapid progress in creative content generation, spanning text, music, and visual art. Yet performative creativity remains notably harder: stand-up comedy is not just “good writing,” but a tightly choreographed sequence of linguistic craft, temporal control, and social-context awareness. This gap is reflected even in industry practice—despite the scale and capability of modern foundation models, major labs rarely report standardized “humor ability,” largely because humor evaluation itself is intrinsically difficult: what counts as funny is subjective, culturally grounded, context-dependent, and highly sensitive to delivery and timing.

\begin{figure}
    \centering
    \includegraphics[width=1\linewidth]{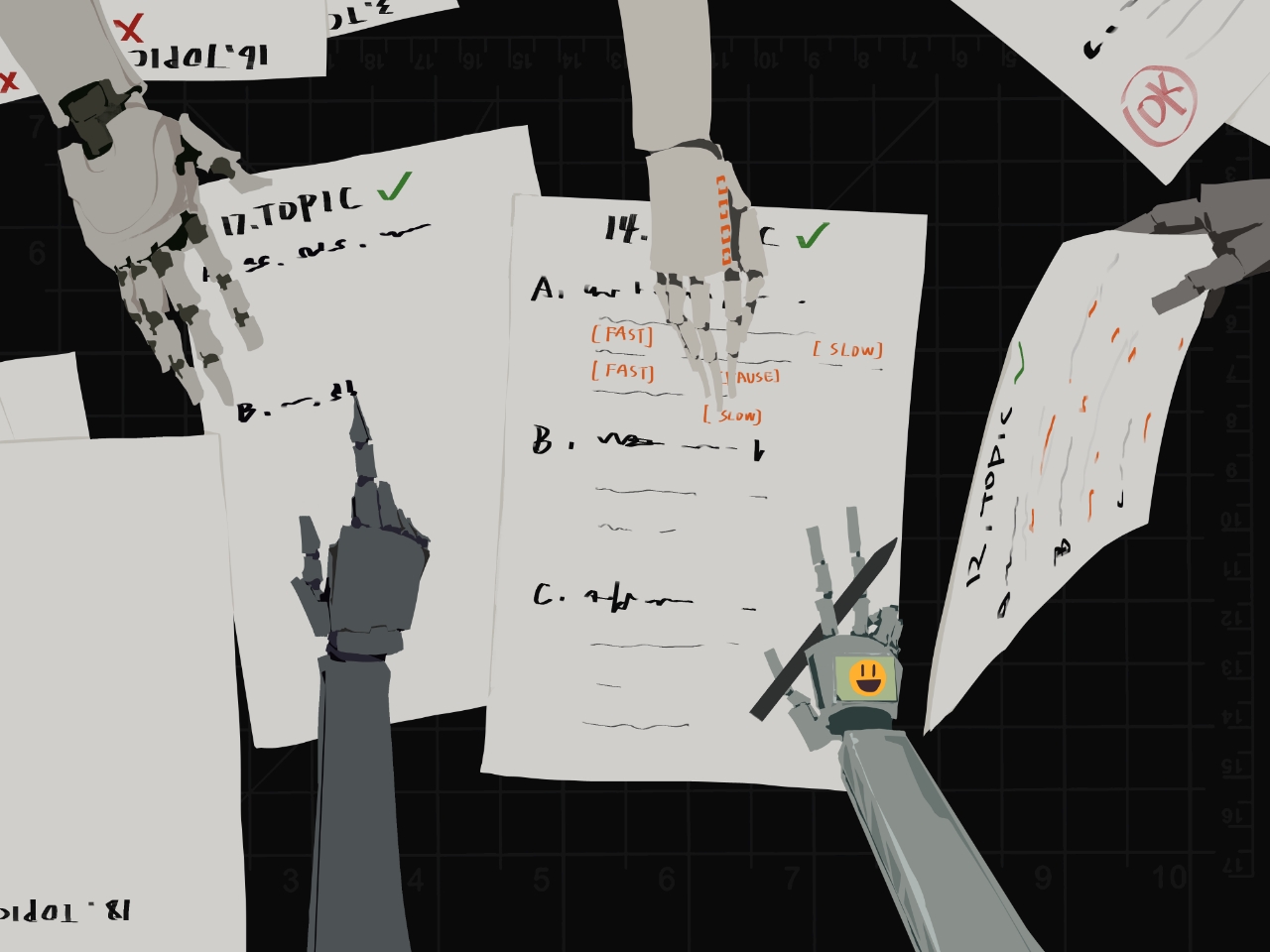}
    \caption{\textbf{Multi-agent collaborative pipeline for Chinese stand-up comedy generation.} Specialized agents iteratively decompose ideation, retrieval, joke writing.}
    \label{fig:first}
\end{figure}
Recent research therefore tends to emphasize humor understanding and evaluation rather than full humor generation, because the former is easier to define and benchmark. In Chinese, this tilt is particularly visible: datasets such as CFunSet \citep{CFUN} provide rich resources for analyzing and probing humor, with tasks including (i) humor cause analysis, (ii) crosstalk “straight-man” response, (iii) humor binary classification, (iv) keyword-based joke generation, (v) topic-conditioned joke generation, and (vi) joke continuation. While these tasks are valuable for modeling comedic signals and building evaluators, they do not directly match the target of long-form Chinese stand-up: a 3–5 minute performance requires coherent comedic arcs, delayed punchlines, callbacks, and stage-ready phrasing—properties that are under-specified by short-form supervision and are hard to learn from understanding-centric labels alone.

Meanwhile, generation remains difficult even with strong general-purpose models. As our preliminary comparison in Fig. \ref{fig:gpt_full} and \ref{fig:deepseek_full} suggests, a strong general model (e.g., GPT-5.2) can drift into didactic or “preachy” narration when asked to produce stand-up, while another strong Chinese model (e.g., DeepSeek) may produce jokes that are sparse and uneven in quality. These failures are not simply stylistic; they reveal missing control over (1) comedic structure (setup–punchline delay, misdirection, callback), (2) timing (pauses, emphasis, rhythm), and (3) performability (spoken language and stage cues). In other words, humor is not equivalent to fluent text, and Chinese stand-up further amplifies the challenge through heavier reliance on shared social context, colloquial delivery, and timing-sensitive audience expectation management.

To address these issues, we build on the intuition that stand-up generation is closer to a production pipeline than a single-shot completion: it requires planning, audience adaptation, writing, coaching, and critique—each with different objectives and failure modes. We therefore propose OpenMic, a multi-agent system implemented with AutoGen (Wu et al., 2024), where specialized agents collaborate in a multi-round iterative loop to refine content toward both comedic quality and stage readiness. To bridge the dataset–task mismatch, we incorporate retrieval-augmented generation (RAG) to ground writing in diverse comedic materials and to expand topic-specific angles, and we fine-tune a dedicated JokeWriter to better internalize stand-up-oriented structures beyond what understanding-focused datasets naturally provide. Finally, OpenMic outputs not only a script but a structured performance representation (e.g., pauses, applause beats, emphasis) that can be rendered into an end-to-end video.

\begin{itemize}
\item We implement an end-to-end multi-agent Chinese stand-up comedy generation system based on AutoGen, from user topic input to a stage-ready performance.
\item We introduce RAG-based material retrieval to enrich content grounding and alleviate sparsity/mismatch of stand-up supervision.
\item We design a multi-round iterative self-improve workflow to improve comedic structure, timing, and performability.
\item We fine-tune a dedicated JokeWriter to better capture setup–punchline delay, callback patterns, and spoken-stage style.
\item We propose a structured performance script interface (pauses, applause, emphasis, etc.) and a pipeline that converts it into narrated comedy video output.
\end{itemize}

\section{Related Works}

The field of computational humor has long been considered an "AI-complete" problem because it requires a deep understanding of semantics, pragmatics, and social context. \citep{R1} Traditional research focused on Incongruity Theory, which posits that humor arises from the sudden resolution of a mismatch between expectations and reality.\citep{R2,R3} In the context of crosstalk and talkshows, this is manifested as the "set-up and punchline" logic (or Baofu in Chinese crosstalk). Early attempts at humor generation were often template-based and lacked the creative "logic jump" required for effective comedy.\citep{R4} Recent work such as "Humor Mechanics: Advancing Humor Generation with Multistep Reasoning" has shifted the focus toward reconstructing these mechanics through data-driven policies.\citep{R5} They demonstrate that humor is not merely a linguistic byproduct but a result of multistep reasoning where the model must distill humor principles—such as wordplay and unexpected twists—from existing datasets to generate novel content rather than just acting as a "stochastic parrot."

Our technical framework draws from three rapidly evolving areas of NLP. First, while traditional Retrieval-Augmented Generation (RAG) was primarily used for fact-checking, it has recently been adapted for creative tasks to inject cultural "memes" and specific comedic styles.\citep{RAGFORCREATIVE} Current trends favor Hybrid Adaptation (similar to Retrieval-Augmented Fine-Tuning or RAFT), which balances the static domain expertise of the model with dynamic, external context.\citep{RAGFINETUNE} Second, the development of Parameter-Efficient Fine-Tuning (PEFT) has moved from LoRA to QLoRA, allowing for the specialization of large language models (LLMs) on high-quality comedic scripts without the prohibitive cost of full retraining.\citep{QLORA} Finally, our architecture utilizes a Multi-Agent System (MAS) to mimic human collaborative creativity. We build upon the foundation of works like HoLLMwood which assigns LLMs to specialized roles such as "Writer," "Editor," and "Actor" to improve narrative coherence.\citep{R2} By following the modular design principles outlined in recent MAS surveys, we create a specialized pipeline where different agents handle distinct stages of the crosstalk generation process.

The current landscape of humor generation is increasingly focusing on multi-dimensional evaluation and cultural specificity. For instance, this paper \citep{Assess} reveals that while modern LLMs can match low-to-mid tier human performance in improvisational Japanese comedy, they often prioritize "Novelty" over "Empathy," leading to a divergence in what machines and humans perceive as funny. Similarly, Guo et al. (2023) highlighted the gap in LLM performance for Chinese crosstalk, where models struggle with the rhythmic cadence and the specific structural requirements of the medium.\citep{CANLLMFUN} Our work contributes to this evolving field by combining a novel multi-LLM agent system with a RAG-based context injector and an agent-specific fine-tuning strategy. By specifically training agents to play individual roles (e.g., the "JokeWriter" agent), we explore whether the synergy of specialized roles and retrieved comedic materials can overcome the empathy-novelty gap identified in recent benchmarks.

\section{Humor and Cognitive Reasoning}

\begin{figure}
    \centering
    \includegraphics[width=1\linewidth]{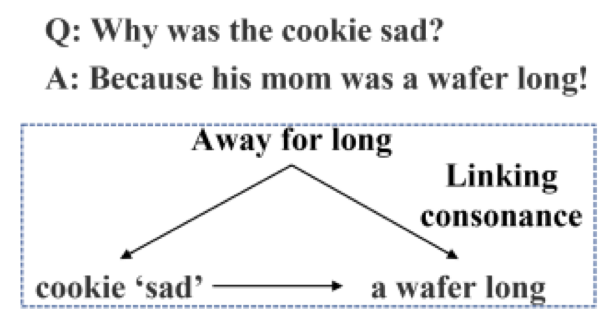}
    \caption{Backdoor Criterion}
    \label{fig:humor_backdoor}
\end{figure}

\begin{figure}
    \centering
    \includegraphics[width=1\linewidth]{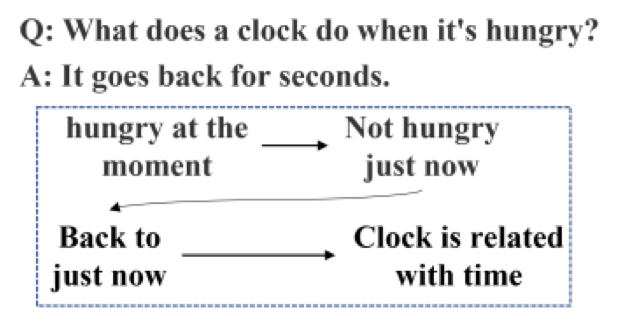}
    \caption{Frontdoor Criterion}
    \label{fig:humor_frontdoor}
\end{figure}

Humor is often treated as a stylistic property of language, yet many jokes are better understood as \emph{reasoning processes} that manipulate an audience’s expectations over time. A stand-up punchline rarely succeeds by lexical novelty alone; rather, it relies on a latent chain of inferences that (i) builds a plausible interpretation, (ii) introduces a hidden connection, and (iii) triggers a rapid ``re-interpretation'' that resolves the incongruity. This view is especially important for Chinese stand-up comedy, where effective jokes frequently depend on shared cultural context, implicit premises, and tightly controlled information release. In this section, we frame humor as a form of structured cognitive reasoning and use two illustrative logic diagrams (Fig.~\ref{fig:humor_backdoor} and Fig.~\ref{fig:humor_frontdoor}) to highlight distinct inference patterns that commonly appear in jokes.

\subsection{Humor as a Form of Cognitive Incongruity}

\label{sec:humor_reasoning}
Classic humor theories emphasize \emph{incongruity}: a joke sets up an expectation and then violates it in a way that still permits a coherent resolution. Under an incongruity--resolution perspective, ``being funny'' is not merely about generating surprising words, but about creating a \emph{controlled mismatch} between the audience’s predicted continuation and the eventual reinterpretation that makes the punchline sensible. Concretely, the setup implicitly constructs a mental model of the situation; the punchline either flips a key assumption or reveals a hidden linkage that forces the audience to update that model. The comedic effect arises from the \emph{contrast} between the initial expectation and the revised interpretation, as well as the speed and clarity with which the resolution becomes apparent.

\subsection{Humor Requires Multi-Step Reasoning}
Many jokes embed a multi-step inference chain rather than a single-step association. Fig.~\ref{fig:humor_backdoor} provides an example we refer to as a \emph{backdoor-style} structure: the question entity $E_Q$ and answer entity $E_A$ are not directly connected by surface meaning, but are linked through an intermediate bridging entity $E_Z$ (often a homophone, pun, or shared attribute). In the illustrated joke (``Why was the cookie sad?''), the surface reading suggests an emotional explanation; the resolution depends on mapping to the phonetic/lexical bridge (e.g., ``away for long'' $\leftrightarrow$ ``a wafer long''), which then retroactively makes the punchline interpretable. Here, $E_Z$ acts as a hidden connector that is easy to miss unless one actively searches for alternative interpretations.

In contrast, Fig.~\ref{fig:humor_frontdoor} illustrates a \emph{frontdoor-style} multi-hop reasoning pattern: the setup encourages the audience to traverse intermediate thoughts explicitly before arriving at the punchline. In the example (``What does a clock do when it's hungry?'' $\rightarrow$ ``It goes back for seconds.''), the humor hinges on composing several simple steps: ``hungry'' evokes a desire for food; ``seconds'' can mean a second helping; and ``clock'' relates to time, enabling the wordplay ``goes back for seconds.'' Compared with the backdoor-style pun, the intermediate entity $E_Z$ in this case is not merely a hidden phonetic bridge but a \emph{conceptual stepping stone} that the listener can traverse through associative and compositional reasoning.

These two patterns are common in stand-up: (i) \textbf{delayed punchlines} resemble multi-step inference with intentionally withheld bridges, and (ii) \textbf{callbacks} resemble long-range reasoning where an earlier premise is reactivated under a new interpretation. As a result, humor quality depends not only on what is said, but on \emph{when} the crucial bridge is revealed and how reliably the audience can reconstruct the implicit reasoning path.

\subsection{Why LLMs Struggle with Humor}
Despite strong general language ability, LLMs frequently underperform on humor because fluent continuation does not guarantee the \emph{cognitive surprise} required for a joke. First, models tend to collapse the inference process: they may reveal the bridge too early, explain the joke explicitly, or smooth over ambiguity---all of which reduce comedic tension. Second, many jokes require maintaining two competing interpretations until the punchline; this demands deliberate control of uncertainty and information release, whereas next-token prediction often favors a single dominant continuation. Third, humor is highly sensitive to pragmatic constraints (social norms, persona, cultural presuppositions), so even logically consistent outputs may fail to land as funny if the implied premises are unnatural for the target audience.

The reasoning structures in Fig.~\ref{fig:humor_backdoor}--\ref{fig:humor_frontdoor} also expose a practical issue: the model must \emph{search} over potential bridges $E_Z$ (phonetic, semantic, or contextual) and then \emph{stage} the reveal at the right moment. Without explicit mechanisms for planning, critique, and timing control, single-shot generation often produces either (i) coherent but unfunny narration, or (ii) isolated one-liners that lack buildup, callbacks, and performance rhythm.

\subsection{Implication: Humor as Structured Reasoning}
Viewing humor as structured reasoning suggests that effective stand-up generation is closer to a pipeline of \textbf{planning}, \textbf{verification}, and \textbf{execution} than to free-form text generation. Planning selects comedic angles and determines which bridge $E_Z$ to hide or surface and when; verification checks whether the reasoning path is reconstructible and whether the punchline resolves the incongruity; execution adds stage-performance cues (pauses, emphasis, and rhythm) that regulate information release. This framing motivates our system design: instead of asking a single model to simultaneously invent content, manage audience adaptation, and control delivery, we decompose the process into specialized roles and iterative refinement so that the final performance preserves both the reasoning structure and the timing-sensitive comedic payoff.

\section{Methodology}

\label{sec:multiagent}
\subsection{Multi-Agent System Design}
\textbf{Why Multi-Agent for Stand-Up Comedy?} As discussed in Sec.~\ref{sec:humor_reasoning}, humor is tightly coupled with \emph{structured reasoning} and \emph{timing-aware information release}. In practice, Chinese stand-up generation involves several competing objectives that are hard to satisfy in a single pass: (i) \textbf{content planning} (selecting angles, building setups, placing callbacks), (ii) \textbf{audience adaptation} (persona, taboo avoidance, cultural priors), (iii) \textbf{performability} (spoken-style wording, rhythm, pauses, emphasis), and (iv) \textbf{quality assurance} (coherence, novelty, safety, ``laugh potential''). A single-agent LLM prompt tends to entangle these goals and often collapses the latent reasoning structure (e.g., explaining the joke too early) or drifts into generic narration.

We therefore adopt a multi-agent design that decomposes the pipeline into specialized roles with explicit responsibilities. This separation yields two practical advantages: (1) \textbf{controllability}—each agent optimizes a well-defined sub-objective with dedicated constraints; and (2) \textbf{iterative refinement}—failures can be localized and corrected (e.g., rewrite a weak punchline without redoing audience profiling), which is essential for timing-sensitive comedic arcs.
\begin{figure}
    \centering
    \includegraphics[width=1\linewidth]{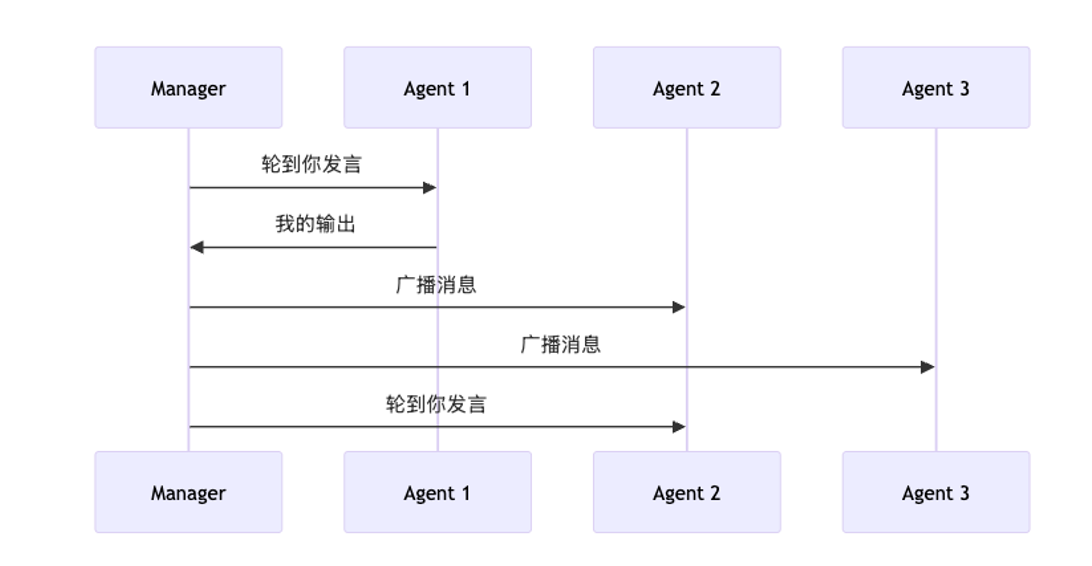}
    \caption{Mechanism of Autogen}
    \label{fig:autogen_mechanism}
\end{figure}
\subsection{AutoGen as the Orchestration Backbone}

We implement OpenMic on top of AutoGen's group-chat abstraction. Conceptually, AutoGen coordinates a set of conversable agents via a manager that (i) decides whose turn it is to speak, (ii) collects the agent's output, and (iii) broadcasts relevant messages to other agents for subsequent turns (Fig.~\ref{fig:autogen_mechanism}). This ``turn-taking + broadcast'' mechanism is a natural fit for creative collaboration: agents can operate asynchronously in intent (each has its own rubric), yet remain synchronized through shared context.

In our implementation, we use a \textbf{GroupChatManager} to enforce an ordered protocol and to prevent uncontrolled multi-agent chatter. Each agent is instantiated as a \textbf{ConversableAgent} with a role-specific system prompt, input/output schema, and constraints. The manager schedules agents following our workflow (Sec.~\ref{sec:workflow}) and handles termination conditions (either \texttt{PASS} from the quality controller or a maximum iteration budget).

\subsection{Blackboard-Centric Coordination}

Beyond message passing, OpenMic employs a \textbf{blackboard} to maintain structured shared state (Fig.~\ref{fig:blackboard_arch}). The blackboard stores intermediate artifacts that must persist across turns and iterations, including:
\begin{itemize}
    \item \textbf{Audience profile}: persona, preferences, taboo list, acceptable language register;
    \item \textbf{Topic expansion}: subtopics, personal anecdotes angles, candidate premises;
    \item \textbf{Draft script}: current version of the stand-up text with section boundaries;
    \item \textbf{Performance markup}: a structured DSL with pauses, emphasis, applause beats;
    \item \textbf{Critique \& action items}: concrete revision instructions and failure reasons.
\end{itemize}
This design prevents critical information from being lost in long conversational context and makes the iteration loop more deterministic: each agent reads from and writes to designated blackboard fields, rather than relying solely on implicit conversational memory.

\subsection{Agent Roles and Interfaces}
\begin{figure}
    \centering
    \includegraphics[width=1\linewidth]{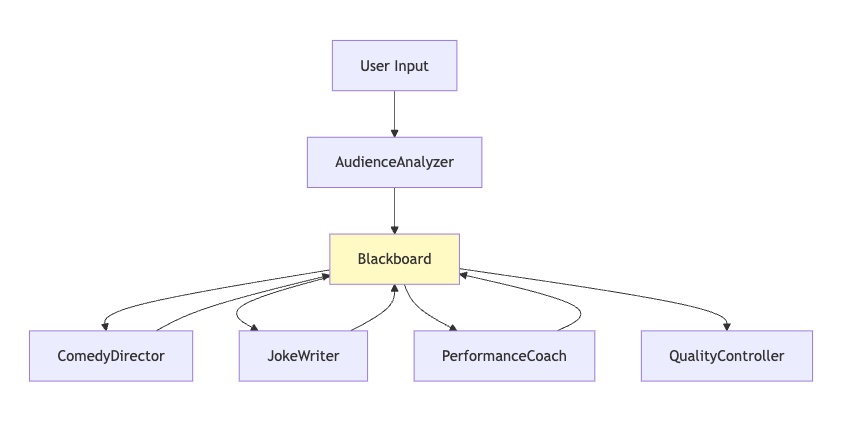}
    \caption{Multi-agent Structure.}
    \label{fig:blackboard_arch}
\end{figure}
OpenMic consists of five core agents (Fig.~\ref{fig:blackboard_arch}), each with a narrowly-scoped responsibility and a typed output that is written to the blackboard:

\paragraph{AudienceAnalyzer (audience modeling).}
Given the user topic and optional style constraints, the AudienceAnalyzer produces an audience-facing \textbf{persona card} and a \textbf{taboo/avoid list} (e.g., sensitive references, overly offensive wording) to ensure cultural and situational appropriateness.

\paragraph{ComedyDirector (high-level planning).}
The ComedyDirector decomposes the topic into a set of \textbf{subtopics} and a \textbf{comedic structure plan} (e.g., opening hook, 2--3 bits, a callback, closing tag). The output is an outline with explicit comedic intent: where tension is built, where bridges are revealed, and where callbacks should land.

\paragraph{JokeWriter (script drafting).}
Conditioned on the outline and audience profile, the JokeWriter produces a \textbf{complete draft script}. We instruct it to maintain spoken Chinese, enforce setup--punchline delays, and preserve long-range dependencies (callbacks) rather than generating disconnected one-liners.

\paragraph{PerformanceCoach (delivery \& markup).}
The PerformanceCoach transforms the draft into a \textbf{performance-ready script} by adding a structured DSL annotation, including pauses (e.g., pre-punchline pause), emphasis, pace changes, filler words, and optional applause/laughter cues. This bridges text generation with downstream audio/video rendering.

\paragraph{QualityController (evaluation \& gating).}
The QualityController acts as a critic and gatekeeper. It evaluates coherence, comedic payoff, timing realism, and audience fit, then outputs either \textbf{PASS} or \textbf{REVISION} with actionable edits. This turns subjective humor quality into an operational criterion for iteration.

\subsection{Hierarchical Multi-Agent RAG with Information Isolation}

Our RAG framework is designed to bridge the gap between simple semantic retrieval and genuine creative transformation. Rather than relying on a traditional single-step retrieval process, we implemented a \textbf{\textcolor{orange}{triadic inner-conversation architecture}} that utilizes one retrieval engine alongside two specialized LLM agents. This system is governed by a custom protocol to ensure that the massive volume of data required for candidate selection does not clutter the primary workflow's context window.

\textbf{Dataset Composition and Post-Processing} 
To ensure stylistic consistency, our retrieval corpus combines two primary sources. The first is a collection of short-form setups and punchlines sourced from the CFUN repository\citep{CFUN}. The second is a  \textbf{\textcolor{orange}{Crosstalk-to-Talkshow Pipeline}} where we took traditional crosstalk scripts and pushed them through an LLM-driven refinement stage. During this process, we performed anonymization by removing specific performer names and executed a stylistic conversion. This turned dialogue-heavy routines into narrative-driven talkshow observations, moving away from the classic "teasing and reacting" dynamic to a more modern first-person perspective.

\textbf{The Triadic Inner-Conversation Workflow} 
Standard semantic matching often prioritizes factual similarity over comedic value. To fix this, we formalize the RAG process as a sequence of three specialized operations: retrieval, scoring, and refinement. 

Let $q$ represent the user topic query and $\mathcal{D}$ the integrated comedic corpus. We define $E(\cdot)$ as the embedding function that maps text to a high-dimensional vector space. The process is defined as follows:

\begin{enumerate}
    \item \textbf{Semantic Retrieval:} The RAG Retriever identifies a set of raw candidates $\mathcal{C}$ by calculating the cosine similarity between the query and document embeddings:
    $$\mathcal{C} = \text{top-}k_{1} \{ d \in \mathcal{D} \mid \text{sim}(E(q), E(d)) \}$$
    where $\text{sim}(\mathbf{u}, \mathbf{v}) = \frac{\mathbf{u} \cdot \mathbf{v}}{\|\mathbf{u}\| \|\mathbf{v}\|}$.

    \item \textbf{LLM Candidate Scoring:} The \textcolor{orange}{LLM Candidate Scorer} (Agent 1) acts as a non-linear semantic filter. It evaluates the comedic potential $P$ of each candidate $c \in \mathcal{C}$ based on latent features like incongruity and relevance, selecting a subset $\mathcal{S}$ of high-potency jokes:
    $$\mathcal{S} = \{ c \in \mathcal{C} \mid f_{\text{scorer}}(c, q) > \tau \}$$
    where $f_{\text{scorer}}$ is the agent's internal evaluation function and $\tau$ is the quality threshold for the top-$k_{2}$ selection.

    \item \textbf{LLM Punchline Refinement:} Finally, the \textbf{LLM Punchline Selector} (Agent 2) performs the creative transformation $\mathcal{T}$. Instead of passing the full text, it distills the selected jokes into a set of writing materials $\mathcal{M}$:
    $$\mathcal{M} = \bigcup_{s \in \mathcal{S}} \mathcal{T}_{\text{selector}}(s)$$
\end{enumerate}

This ensures the JokeWriter receives a distilled set of high-potency building blocks $\mathcal{M}$ rather than a wall of raw, unorganized text, significantly reducing context noise while maximizing creative signal.

\textbf{The "Secret Blackboard" and Context Management}
A key technical feature of our architecture is the  \textbf{\textcolor{orange}{Secret Blackboard}}. During the inner-conversation between the RAG engine and the retrieval agents, thousands of tokens of raw material are processed simultaneously. Storing this in the main global blackboard would quickly exceed the context limits of agents further down the line, such as the Performance Coach. To solve this, the Secret Blackboard acts as a private memory buffer. It only releases the final refined punchlines to the JokeWriter, effectively hiding the noisy retrieval process from the rest of the chain and maintaining a high signal-to-noise ratio across the entire system.

\subsection{Multi-Round Refinement}
\label{sec:workflow}
If the QualityController returns \texttt{REVISION}, the system re-enters the loop by routing feedback back to the JokeWriter (and optionally the PerformanceCoach) until either \texttt{PASS} is obtained or a maximum number of rounds is reached. This multi-round loop is crucial for stand-up: jokes often fail due to localized issues (weak punchline, premature explanation, missing callback trigger, unnatural pause placement) that are best fixed through targeted rewrites rather than regenerating everything from scratch.

\paragraph{Dual-Dimension Quality Assessment}
Unlike monolithic QA systems, our \texttt{QualityController} performs \textbf{dual-dimension evaluation} $Q_r = (Q_r^R, Q_r^W)$ to separately assess retrieval quality and writer quality.

\textbf{RAG Dimension} ($Q_r^R$): Evaluates retrieved joke material quality—humor potential, topic relevance, and diversity. When $Q_r^R = 0$, the QA outputs:

\begin{itemize}[leftmargin=*,noitemsep,topsep=0pt]
    \item $\mathbf{k}^*$: refined keywords
    \item $\mathcal{E}_r$: joke IDs to exclude in next retrieval
    \item $f_r^R$: specific feedback per joke
\end{itemize}

\textbf{Writer Dimension} ($Q_r^W$): Evaluates script organization via three checks:
\begin{equation}
Q_r^W = \mathbb{1}[\text{struct}] \land \mathbb{1}[\text{safe}] \land \mathbb{1}[\text{length}]
\end{equation}
Failed checks trigger \textit{rewrite directives} $\mathbf{d}_r$ (e.g., ``callback missing for setup in line 3'').

\paragraph{Targeted Refinement Routing}
The dual evaluation enables surgical fixes:

\textit{Case 1: RAG fails, Writer succeeds} ($Q_r^R=0, Q_r^W=1$):
\begin{equation}
\text{Re-retrieve: } \mathcal{D}_{r+1} \leftarrow \texttt{RAG}(\mathbf{k}^*, \mathcal{E}_r)
\end{equation}

\textit{Case 2: Writer fails, RAG succeeds} ($Q_r^R=1, Q_r^W=0$):
\begin{equation}
\text{Rewrite: } s_{r+1} \leftarrow \texttt{Writer}(\mathcal{D}_r, \mathbf{d}_r)
\end{equation}

\textit{Case 3: Both fail} ($Q_r^R=0, Q_r^W=0$):
\begin{equation}
\mathcal{D}_{r+1}, s_{r+1} \leftarrow \texttt{RAG}(\mathbf{k}^*, \mathcal{E}_r) + \texttt{Writer}(\cdot, \mathbf{d}_r)
\end{equation}

Termination occurs when $Q_r^R \land Q_r^W = 1$ or $r \geq R_{\max}$.

\paragraph{Context-Aware Memory}
Each writer receives structured feedback:
\begin{align}
\mathcal{C}_r = \{s_{r-1}, &\mathcal{D}_{r-1}, \mathbf{d}_{r-1}, \nonumber\\
&Q_{r-1}^R, \mathcal{P}_{r-1}\}
\end{align}
where $\mathcal{P}_{r-1}$ are preserved joke IDs. This prevents agents from ``forgetting'' prior decisions across rounds.

Empirically, Case 3 occurs in $\sim$30\% of round-1 attempts but drops to $<$5\% by round 3, indicating rapid convergence.

\subsection{Domain-Specific Adaptation via QLoRA}
To bridge the gap between general-language capabilities and specialized comedic timing, we employ \textbf{Quantized Low-Rank Adaptation (QLoRA)}. This approach allows for the fine-tuning of large-scale models by injecting trainable low-rank matrices into the frozen, 4-bit quantized base model. For a weight matrix $W_0 \in \mathbb{R}^{d \times k}$, the forward pass is modified as:

$$h = W_0 x + \Delta W x = W_0 x + BAx$$

where $B \in \mathbb{R}^{d \times r}$ and $A \in \mathbb{R}^{r \times k}$ are the low-rank adapters with rank $r \ll \min(d, k)$. We specifically target all \textcolor{orange}{linear projections} within the transformer blocks to maximize the model's stylistic plasticity. Furthermore, to ensure the model focuses exclusively on comedic delivery, we utilize a \textcolor{orange}{completion-only loss strategy}, calculating gradients only on the generated punchlines rather than the instruction prompts.

\section{Experiments}

\subsection{LLM-as-a-Judge Evaluation Framework}We implemented a rigorous "LLM-as-a-Judge" mechanism to quantify the quality of the generated talk show scripts. Unlike standard NLP metrics (such as BLEU or ROUGE), which often fail to capture the semantic nuance and comedic timing of creative writing, we utilized a senior executive producer persona—powered by the \texttt{Grok-4-1-fast-reasoning} model—to conduct a multi-dimensional scoring analysis.The evaluation is governed by a Pydantic-enforced schema, ensuring that every assessment is structured across five critical dimensions:\begin{itemize}\item \textbf{Persona Fidelity (30\%):} The distinctiveness and consistency of the characters' voices.\item \textbf{Humor Mechanics (25\%):} The density and structural quality of setup-punchline sequences.\item \textbf{Interactive Reactivity (20\%):} The degree of "improvisational" riffing and response to previous turns.\item \textbf{Contextual Coherence (15\%):} The logical consistency and effective use of callbacks.\item \textbf{Narrative Arc (10\%):} The rhythmic flow from introduction to climax and resolution.\end{itemize}The final weighted score $S_{total}$ is calculated as:

$$S_{total} = 0.30P + 0.25H + 0.20R + 0.15C + 0.10N$$

where $P, H, R, C, N$ represent the scores for Persona, Humor, Reactivity, Coherence, and Narrative respectively.

\begin{table}[t]
\centering
\small
\caption{Evaluation Results across Different Temperature Settings}
\label{tab:temp_results}
\begin{tabular}{lccccc}
\toprule
\textbf{Configuration} & 
\rotatebox{75}{\textbf{Persona}} & 
\rotatebox{75}{\textbf{Reactivity}} & 
\rotatebox{75}{\textbf{Humor}} & 
\rotatebox{75}{\textbf{Narrative}} & 
\rotatebox{75}{\textbf{Coherence}} \\
\midrule
JW+Tem0.1 & 82.5 & 88.0 & 95.5 & 93.0 & 97.5 \\
JW+Tem0.3 & 88.5 & 15.0 & 96.0 & 93.0 & 97.5 \\
JW+Tem0.5 & 82.5 & 15.0 & 96.5 & 93.0 & 98.5 \\
JW+Tem0.7 & 92.5 & 45.0 & 96.0 & 98.5 & 97.0 \\
JW+Tem0.9 & 85.0 & 25.0 & 92.0 & 94.0 & 96.0 \\
\bottomrule
\end{tabular}
\end{table}

\begin{figure}
    \centering
    \includegraphics[width=1.0\linewidth]{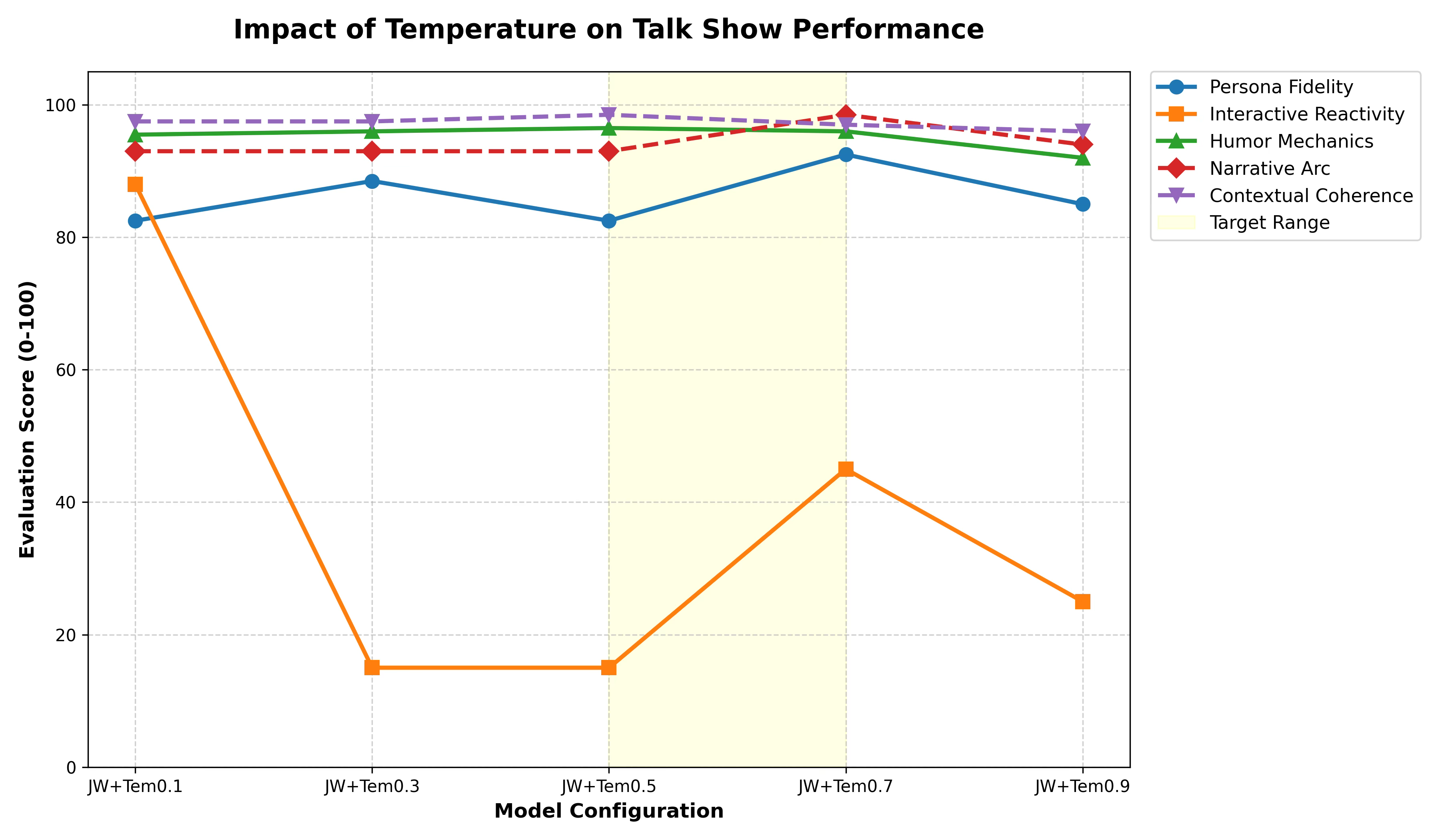}
    \caption{visualization of generation scores under different temperature paramters}
    \label{fig:temp}
\end{figure}

\subsection{Influence of Temperature on Generative Performance}

While initial metrics exhibited a high degree of variance, they provided critical insights into the relationship between sampling temperature and the efficacy of our RAG-enhanced retrieval. We observed that the system's performance is not a linear function of stochasticity, but rather a delicate balance between \textcolor{orange}{contextual grounding} and creative divergence.

Our analysis indicates that \textbf{low temperature settings (e.g., $T=0.1$)} combined with the \textbf{large joke corpus retrieved via RAG} yield the most superior results. As shown in \ref{fig:1}, which compares two specific generative examples, lower temperatures allow the model to maintain a high "focus" on the specific comedic building blocks provided by the RAG inner-conversation. In this regime, the \textbf{Contextual Coherence} (97.5) and \textbf{Interactive Reactivity} (88.0) are maximized, as the model accurately maps the retrieved punchlines onto the target persona without drifting into irrelevant hallucinations.

Conversely, at \textbf{higher temperatures (e.g., $T \geq 0.7$)}, the model begins to lose its grip on the retrieved context. While this occasionally results in a spike in \textbf{Narrative Arc} (98.5 at $T=0.7$) as the model explores more varied sentence structures, it frequently compromises \textbf{Reactivity}. Qualitative review suggests that at high temperatures, the JokeWriter often ignores the specific "setup" provided by the RAG selector in favor of generic, less interesting tropes. 

Ultimately, we conclude that the optimal configuration for generative crosstalk lies in \textcolor{orange}{minimizing entropy to maximize retrieval signal}. By utilizing a low temperature, we ensure that the fine-tuned model acts as a precision instrument that "assembles" the retrieved comedic materials into a cohesive script, rather than attempting to hallucinate humor without sufficient grounding. This reinforces the value of our RAG-centric approach: the "creativity" is supplied by the diversity of the corpus, while the "logic" is preserved by the constrained sampling.

\subsection{Finetuning implementation Details and Hyperparameters}
Our fine-tuning experiments were conducted on a single GPU using the \texttt{trl} and \texttt{peft} libraries. The training corpus consists of the LLM-processed Talkshow dataset, formatted using the Qwen-2.5 chat template.

\textbf{Fine-tuning Configuration:} 
We utilized the \texttt{LoraConfig} to target a comprehensive set of modules, including \textcolor{orange}{q\_proj, k\_proj, v\_proj, o\_proj}, and the MLP layers (\textcolor{orange}{gate, up, down\_proj}). The detailed hyperparameter settings are summarized in Table~\ref{tab:hypers}.

\begin{table}[h]
\centering
\caption{Hyperparameters for Comedy-Specialized QLoRA Fine-tuning}
\label{tab:hypers}
\begin{tabular}{ll}
\hline
\textbf{Hyperparameter} & \textbf{Value} \\ \hline
Base Model & Qwen-2.5-3B-Instruct \\
Quantization & 4-bit NF4 (NormalFloat) \\
LoRA Rank ($r$) & 16 \\
LoRA Alpha ($\alpha$) & 32 \\
LoRA Dropout & 0.05 \\
Learning Rate & $2 \times 10^{-4}$ \\
Optimizer & Paged AdamW 32-bit \\
Batch Size (Per Device) & 2 \\
Gradient Accumulation & 4 \\
Training Epochs & 1 \\
Compute Precision & BF16 (or FP16) \\ \hline
\end{tabular}
\end{table}

\textbf{Completion-Only Training:} 
To prevent the model from overfitting on the instruction syntax, we implemented a \textcolor{orange}{ManualCompletionCollator}. By defining the response template as \texttt{"<|im\_start|>assistant\textbackslash n"}, the trainer effectively masks the prompt tokens during loss calculation. This ensures the negative log-likelihood loss $\mathcal{L}$ is computed only on the tokens $y_i$ belonging to the assistant's response:

$$\mathcal{L} = - \sum_{i \in \text{Response}} \log P(y_i \mid y_{<i}, \text{Prompt})$$

The final model was deployed via a vLLM entrypoint with LoRA support enabled, allowing for high-throughput inference during the multi-agent execution phase.

\subsection{Demo: End-to-End Stand-Up Generation}

\paragraph{Goal.}
We present an end-to-end demo to illustrate how our multi-agent pipeline generates a Chinese stand-up monologue from a high-level prompt, highlighting (i) structured setup--punchline planning, (ii) iterative critique and rewriting, and (iii) callback triggering across the script.

\paragraph{Setup.}
We run the system for \emph{3} iterations producing intermediate artifacts including \emph{Quality Evaluations}.

\begin{figure*}[t]
    \centering
    \includegraphics[width=\textwidth]{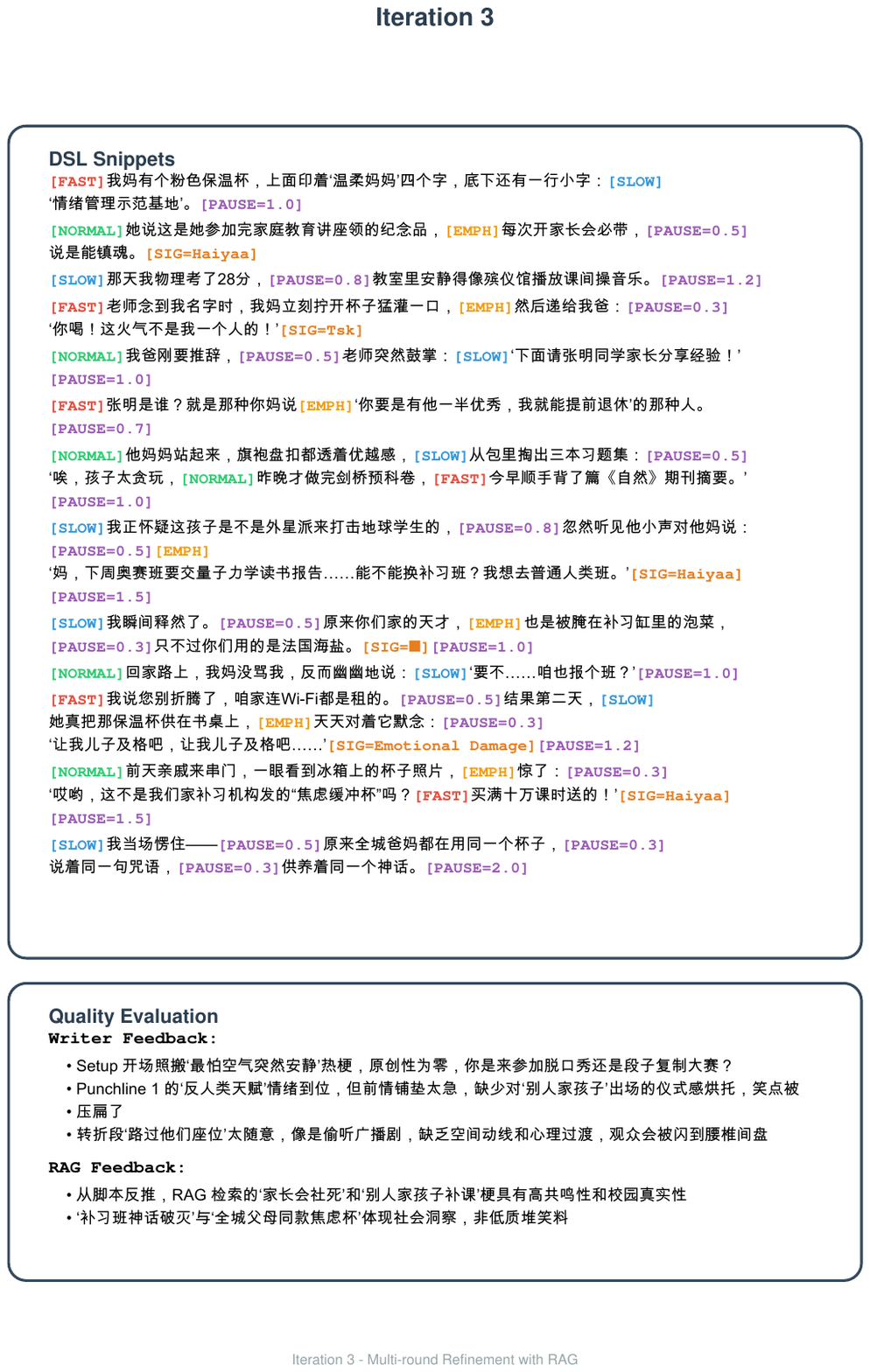}
    \label{fig:demo}
\end{figure*}

\subsection{Downstream Application: End-to-End Video Synthesis}
\label{subsec:video_synthesis}

To further demonstrate the practical utility of \textsc{OpenMic}, we extend the pipeline to a multi-modal application stage. This stage verifies that the structured performance scripts, enriched with behavioral cues, can be seamlessly executed by external rendering engines to produce broadcast-ready content.

\textbf{Implementation Workflow:} The video synthesis process acts as a specialized consumer of the \textit{PerformanceCoach}'s output. We implement a middleware that parses the embedded DSL markers---such as \texttt{[pause]}, \texttt{[emphasis]}, and \texttt{[applause]}---to construct a synchronized temporal timeline. By invoking RESTful APIs from high-fidelity digital human platforms (e.g., Kling AI), the system maps the synthesized audio onto a 3D-animated avatar. The synchronization logic ensures that the avatar’s micro-expressions, such as eyebrow movements during a setup and a smirk during a punchline, are aligned with the comedic rhythm defined in the script.

\textbf{Key Technical Challenges and Observations:}
\begin{itemize}
    \item \textbf{Temporal Consistency:} The use of structured markers prevents the common ``robotic delivery'' seen in standard Text-to-Speech (T2S) systems. By explicitly injecting silence durations and speech rate variations based on the DSL, we preserve the timing-sensitive nature of Chinese stand-up comedy.
    \item \textbf{Cross-Modal Stylistic Alignment:} The visual persona, including stage background illumination and character attire, is dynamically selected to match the \textit{AudienceAnalyzer}'s persona card. This ensures a coherent comedic atmosphere where the visual environment reinforces the linguistic tone.
    \item \textbf{Performance Fidelity:} Our pipeline automates the generation of a 3--5 minute narrated video from a single topic prompt. This end-to-end capability demonstrates the robustness of \textsc{OpenMic} not only as a writing assistant but as a comprehensive production tool for digital entertainment.
\end{itemize}

The integration of video synthesis completes the generative loop, providing a tangible interface for evaluating the performability of the generated humor in a real-world setting.

\section*{Acknowledgments}
This project is a course final assignment for the CoRE course. 
It was developed by the group ``King of Comedy''.

\bibliography{example_paper}
\bibliographystyle{icml2026}

\newpage
\appendix
\onecolumn

\section{Single Agent Example}

\begin{figure*}[t]
    \centering
    \includegraphics[width=0.80\textwidth]{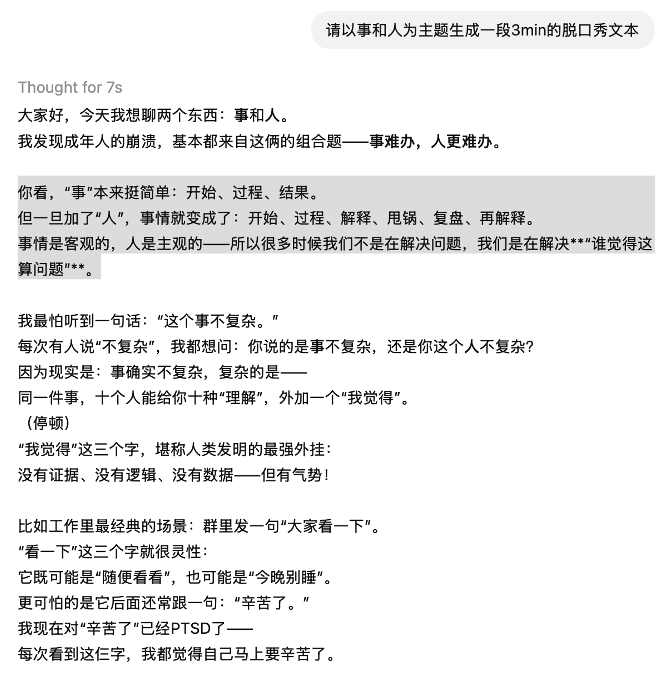}
    \caption{\textbf{GPT}. Qualitative example of Chinese stand-up generation on the same topic prompt.}
    \label{fig:gpt_full}
\end{figure*}

\begin{figure*}[t]
    \centering
    \includegraphics[width=0.90\textwidth]{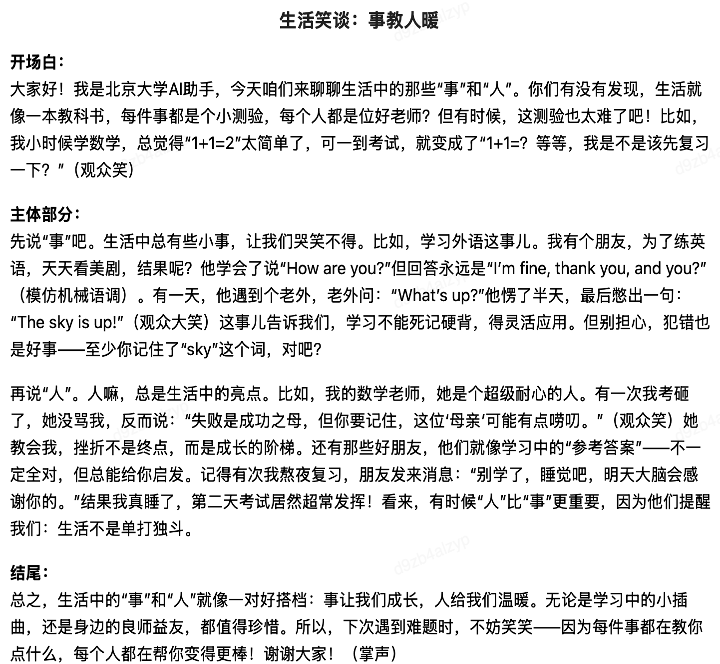}
    \caption{\textbf{DeepSeek}. Qualitative example of Chinese stand-up generation on the same topic prompt.}
    \label{fig:deepseek_full}
\end{figure*}

\section{Single Finetuned Agent Example}

\begin{figure*}
    \centering
    \includegraphics[width=0.9\linewidth]{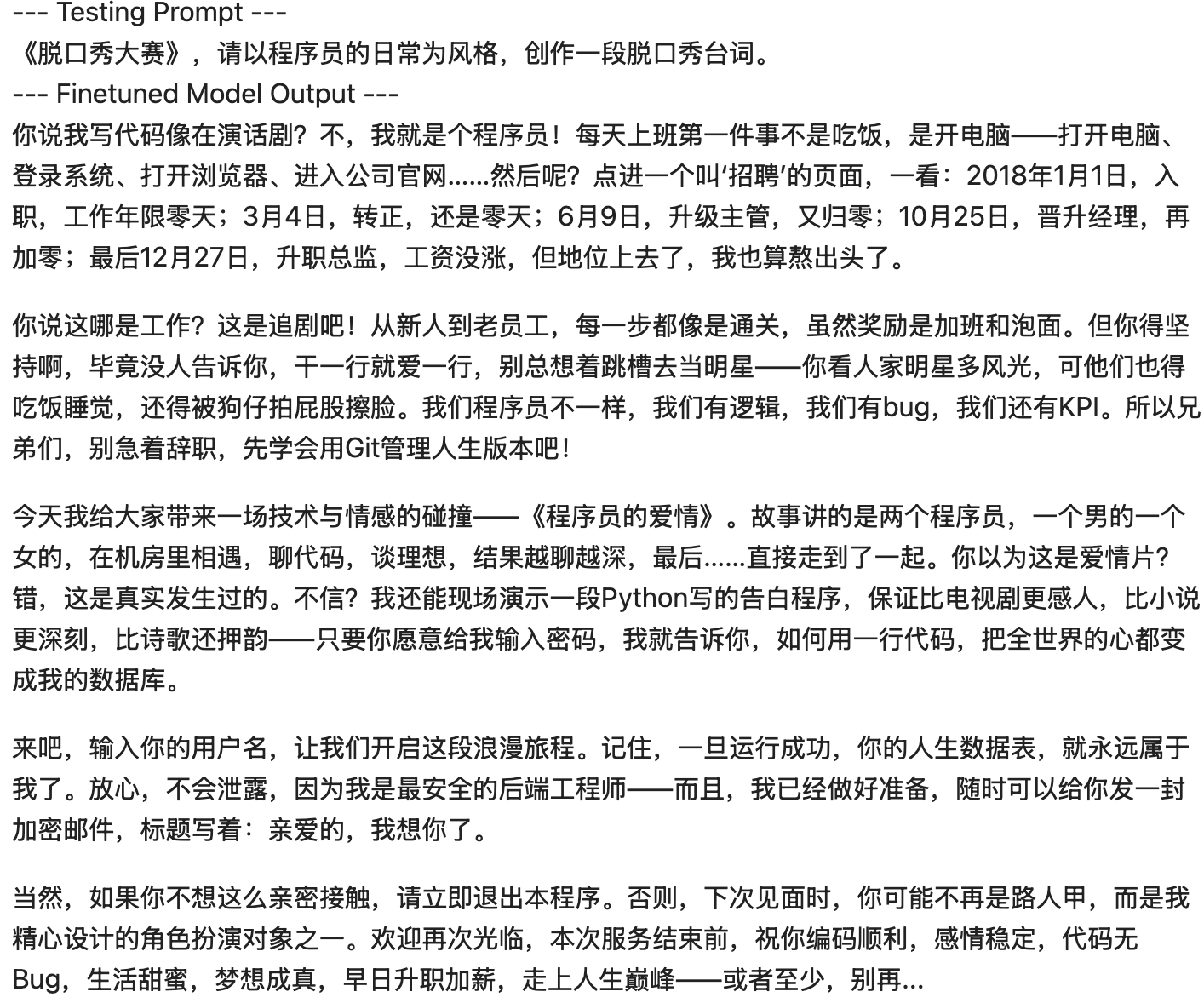}
    \caption{QLora finetuned Qwen3-4B-Instruct}
    \label{fig:finetuned}
\end{figure*}

\section{Different Temperature Setting Examples}

\begin{figure*}
    \centering
    \includegraphics[width=1.0\linewidth]{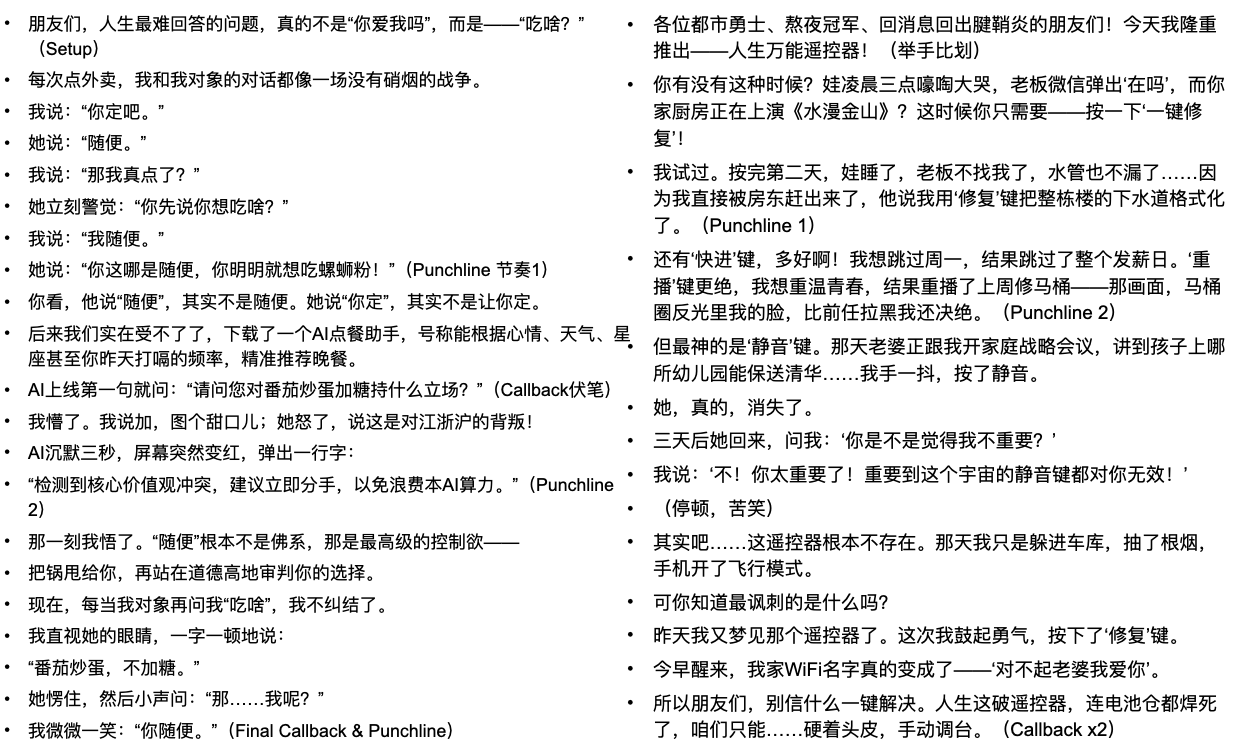}
    \caption{Left one generated with 0.1 temperature, right one with 0.9 temperature}
    \label{fig:1}
\end{figure*}

\end{document}